\documentclass{article}
\usepackage{iftex}
\ifPDFTeX
  \usepackage[T1]{fontenc}
  \usepackage{times}
\else
  \usepackage{fontspec}
\fi

\usepackage[letterpaper,textwidth=5.5in,textheight=9in,top=0.78in]{geometry}
\usepackage{natbib}
\setcitestyle{authoryear,round,citesep={;},aysep={,},yysep={;}}
\usepackage{titlesec}
\usepackage{fancyhdr}
\usepackage{amsmath,amssymb}
\usepackage{booktabs}
\usepackage{array}
\usepackage{listings}
\usepackage{graphicx}
\usepackage{multirow}
\usepackage{xcolor}
\usepackage{caption}
\usepackage{placeins}
\usepackage{float}
\usepackage{url}
\usepackage{hyperref}
\hypersetup{
  hidelinks,
  pdftitle={Harness Annealing: Learning to Act with Less External Control},
  pdfauthor={Yingxuan Yang, Huacan Chai, Ying Wen}
}
\usepackage{microtype}
\usepackage{tikz}
\usetikzlibrary{arrows.meta,calc,positioning}
\usepackage{enumitem}

\newcommand{\method}{\textsc{Harness Annealing Training}}

\definecolor{accentred}{HTML}{A52A2A}
\definecolor{modelblue}{HTML}{DCEBFA}
\definecolor{fixedgray}{HTML}{ECEBE6}
\definecolor{tablered}{HTML}{F4E4E2}
\definecolor{tableblue}{HTML}{E7F0F8}
\definecolor{tablegray}{HTML}{F2F2F0}
\newcommand{\frontiercell}[1]{\colorbox{tablered}{\strut#1}}

\definecolor{codegray}{RGB}{248,248,248}
\definecolor{promptblue}{HTML}{2F5D7C}
\definecolor{promptborder}{HTML}{AFC0CC}
\definecolor{promptbg}{HTML}{F7F9FB}
\definecolor{promptmid}{HTML}{6F7377}
\lstdefinestyle{judgeprompt}{
  basicstyle=\ttfamily\fontsize{7.4}{9.1}\selectfont,
  backgroundcolor=\color{promptbg},
  frame=single,
  rulecolor=\color{promptborder},
  framerule=0.5pt,
  framesep=5pt,
  xleftmargin=0.35em,
  xrightmargin=0.35em,
  framexleftmargin=0.45em,
  framexrightmargin=0.45em,
  breaklines=true,
  breakatwhitespace=true,
  columns=fullflexible,
  keepspaces=true,
  showstringspaces=false,
  emph={CRITICAL,EVALUATION,PRINCIPLES,INPUT,OUTPUT,SCORING,INSTRUCTIONS,REQUIREMENT,Correctness,Completeness,Relevance,Clarity,Reasoning},
  emphstyle=\color{promptblue}\bfseries,
  literate={—}{{\textemdash}}1,
  aboveskip=0.8em,
  belowskip=0.8em
}

\usepackage{colortbl}
\usepackage{tabularx}
\usepackage{longtable}
\usepackage{needspace}
\usepackage{tcolorbox}
\tcbuselibrary{skins,breakable,listings}

\definecolor{appendixInk}{HTML}{344D60}
\definecolor{appendixRule}{HTML}{AAB8C2}
\definecolor{appendixHeader}{HTML}{E8EFF4}
\definecolor{appendixWash}{HTML}{F3F6F8}
\definecolor{appendixPanel}{HTML}{FAFBFC}
\definecolor{appendixMuted}{HTML}{78858F}

\newcommand{\appendixTableStyle}{%
  \small
  \renewcommand{\arraystretch}{1.27}%
  \setlength{\tabcolsep}{6pt}%
  \setlength{\aboverulesep}{0pt}%
  \setlength{\belowrulesep}{0pt}%
  \arrayrulecolor{appendixRule}%
}
\newcommand{\appendixTopRule}{\arrayrulecolor{appendixInk}\toprule[1.05pt]\arrayrulecolor{appendixRule}}
\newcommand{\appendixBottomRule}{\arrayrulecolor{appendixInk}\bottomrule[1.05pt]\arrayrulecolor{appendixRule}}
\newcommand{\apYes}{\textcolor{appendixInk}{$\checkmark$}}
\newcommand{\apNo}{\textcolor{appendixMuted}{--}}

\lstdefinestyle{appendixjudge}{
  basicstyle=\ttfamily\fontsize{8}{9.5}\selectfont,
  frame=none,
  breaklines=true,
  breakatwhitespace=true,
  breakautoindent=true,
  breakindent=1em,
  columns=fullflexible,
  keepspaces=true,
  showstringspaces=false,
  xleftmargin=0pt,
  xrightmargin=0pt,
  aboveskip=0pt,
  belowskip=0pt,
  emph={CRITICAL,EVALUATION,PRINCIPLES,INPUT,OUTPUT,SCORING,INSTRUCTIONS,REQUIREMENT,Correctness,Completeness,Relevance,Clarity,Reasoning},
  emphstyle=\color{appendixInk}\bfseries,
  literate={—}{{\textemdash}}1
}
\newtcblisting{judgepromptbox}[1][Verbatim judge prompt]{
  enhanced,
  breakable,
  listing only,
  listing engine=listings,
  listing options={style=appendixjudge},
  title={#1},
  title after break={Verbatim judge prompt (continued)},
  fonttitle=\bfseries\small,
  coltitle=appendixInk,
  colbacktitle=appendixHeader,
  colback=appendixPanel,
  colframe=appendixRule,
  boxrule=0.5pt,
  titlerule=0.4pt,
  arc=0pt,
  outer arc=0pt,
  width=\linewidth,
  boxsep=0pt,
  left=9pt,
  right=9pt,
  top=8pt,
  bottom=8pt,
  lefttitle=9pt,
  righttitle=9pt,
  toptitle=6pt,
  bottomtitle=6pt,
  before skip=8pt,
  after skip=10pt,
  pad at break*=2mm
}

\newtcolorbox{metricdefinition}[2]{
  enhanced,
  title={#1\hfill\normalfont #2},
  fonttitle=\bfseries\small,
  coltitle=appendixInk,
  colbacktitle=appendixWash,
  colback=white,
  colframe=appendixRule,
  boxrule=0.6pt,
  titlerule=0.3pt,
  arc=0pt,
  outer arc=0pt,
  boxsep=0pt,
  left=8pt,right=8pt,top=4pt,bottom=4pt,
  lefttitle=8pt,righttitle=8pt,toptitle=3pt,bottomtitle=3pt,
  before skip=6pt,after skip=6pt,
  before upper={%
    \setlength{\abovedisplayskip}{6pt}%
    \setlength{\belowdisplayskip}{6pt}%
    \setlength{\abovedisplayshortskip}{3pt}%
    \setlength{\belowdisplayshortskip}{3pt}%
  },
  width=\linewidth
}

\renewcommand{\headrulewidth}{0.4pt}
\titleformat{\section}{\large\scshape\raggedright}{\thesection}{0.75em}{}
\titlespacing*{\section}{0pt}{2.0ex}{1.5ex}
\titleformat{\subsection}{\normalsize\scshape\raggedright}{\thesubsection}{0.75em}{}
\titlespacing*{\subsection}{0pt}{1.8ex}{0.8ex}
\titleformat{\subsubsection}{\normalsize\scshape\raggedright}{\thesubsubsection}{0.75em}{}
\titlespacing*{\subsubsection}{0pt}{1.5ex}{0.5ex}

\renewenvironment{abstract}{%
  \vspace{0.05in}\begin{center}{\large\scshape Abstract}\end{center}%
  \vspace{0ex}\begin{center}\begin{minipage}{4.5in}\noindent%
}{%
  \end{minipage}\end{center}\vspace{1ex}%
}

\begin{document}
\vspace*{0.01in}
{\LARGE\scshape\raggedright
Harness Annealing: Learning to Act with Less External Control\par}
\vspace{0.2in}

{\normalsize\bfseries
Yingxuan Yang\textsuperscript{1,2,*}\qquad
Huacan Chai\textsuperscript{1,*}\qquad
Ying Wen\textsuperscript{1,\ensuremath{\dagger}}\par}
\vspace{0.06in}

{\small
\textsuperscript{1}Shanghai Jiao Tong University \quad
\textsuperscript{2}University of California, Berkeley \par}
\vspace{0.20in}

\begingroup
\renewcommand{\thefootnote}{\fnsymbol{footnote}}
\footnotetext[1]{These authors contributed equally.}
\footnotetext[2]{Corresponding author: Ying Wen.}
\endgroup

\begin{abstract}
Language agents rely on external harnesses to track state, organize workflows, and verify answers.
Beyond providing tools and information, these harnesses supply control decisions about what to investigate, whether to revise, and when to stop.
Training on successful harness-supported trajectories can improve task performance while leaving these decisions dependent on runtime intervention.
We ask whether harness-supported experience can also teach the model to make these decisions, allowing the division of control to change as the model learns.
We call this objective \textbf{\emph{harness internalization}}: learning to assume specified control responsibilities while retaining task performance after the corresponding support is withdrawn.
We introduce \textbf{\method{} (HAT)}, which combines explicit control supervision with a curriculum over teacher trajectories collected under progressively weaker harnesses.
Experiments with 9B and 35B models on SWE-QA and SWE-QA-Pro evaluate every checkpoint under four deployment harnesses.
Selected annealed checkpoints operating with tools alone achieve scores close to those of their respective starting checkpoints deployed with the full harness.
The benefits vary with model scale and deployment configuration, and further annealing does not uniformly improve performance.
These findings suggest that harness-supported experience can help reduce the runtime control required by a trained agent.
\end{abstract}

\section{Introduction}
\label{sec:intro}

LLM agents typically consist of a backbone model and an external execution framework, commonly referred to as a \emph{harness} \citep{yang2024sweagent,zhang2026stopcomparing}.
The harness supports task execution by maintaining state, organizing workflows, and checking candidate answers \citep{chen2026harnessx}.
These components supply control decisions about what to investigate, whether further evidence is needed, and when to stop.
Successful harness-supported interactions therefore provide experience from which a model can learn both task execution and the judgments that guide it.
However, training on these interactions can improve task performance while leaving the model dependent on external intervention for those judgments.
The model may thus continue to require the same runtime guidance even after learning from the experience it helped produce.
If these control responsibilities can be learned, support that was useful for generating training experience may become unnecessary at deployment.
This motivates a central question: \textbf{\emph{can harness-supported experience teach the model to assume specified control responsibilities while preserving task performance as the corresponding support is withdrawn?}}

\begin{figure*}[t]
  \centering
  \includegraphics[width=0.99\textwidth]{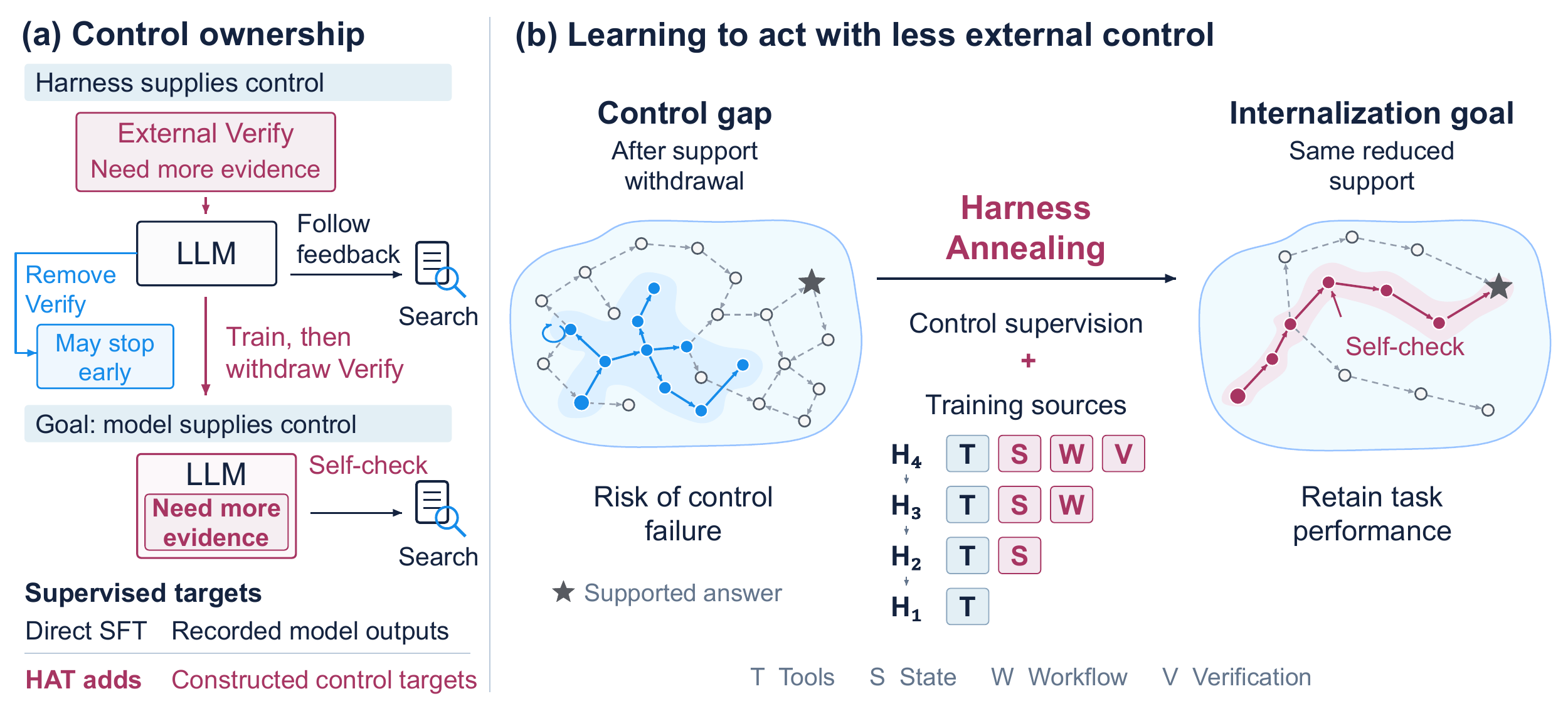}
  \vspace{-0.2cm}
    \caption{\textbf{Learning to retain task performance with less external control.}
    \textbf{(a) Control ownership.} Following verifier feedback does not ensure
    independent judgment after its removal; state, workflow, and tools remain.
    \textbf{(b) Harness annealing.} HAT combines explicit control supervision
    with a curriculum over progressively weaker source harnesses.
    Paths schematically contrast potential control gaps with intended behavior
    under the same reduced support.}
  \label{fig:concept}
  \vspace{-0.1cm}
\end{figure*}

We call the corresponding learning objective \emph{\textbf{Harness Internalization}}: learning to assume specified control responsibilities while maintaining task performance after the associated external support is withdrawn.
We focus on state tracking, workflow organization, and answer verification; tools and execution infrastructure remain external.
Figure~\ref{fig:concept} illustrates this objective through verification: after external checking is removed, the model must assess whether the available evidence supports its answer and continue exploration when necessary.
However, continued exploration in a successful trajectory may simply be a response to a verifier rejecting the model's draft.
Learning the actions that follow this intervention does not guarantee that the model will independently recognize the need for further investigation.
Successful execution under the full harness is therefore insufficient to establish internalization.
Evaluation must also examine task performance and control-related behavior after the corresponding support is removed.

Existing work improves model--harness collaboration in several ways.
Adaptive harnesses optimize workflows, runtime code, or composable components to improve external support \citep{lee2026metaharness,chen2026harnessx}.
Trajectory-based training instead updates the model using successful interactions \citep{chen2023fireact,zeng2023agenttuning}, but harness-generated prompts, states, and interventions typically remain conditioning inputs rather than explicit supervision targets.
Related approaches, including skill distillation and workflow compilation, explore transferring external guidance into model weights.
Our focus is the transfer of specified runtime control responsibilities, assessed through task performance and control-related behavior after support withdrawal.
We study this transfer through a hierarchy of state, workflow, and verification support, evaluating each checkpoint both with and without the corresponding control layers.

We introduce \textbf{\method{} (HAT)}, which combines explicit control supervision with a stronger-to-weaker harness curriculum.
Using a fixed bank of teacher trajectories, we retain interaction observations and teacher actions while constructing control annotations that distinguish externally supplied information from judgments the model must learn to produce.
Starting from a checkpoint trained on full-harness trajectories, training progressively introduces records from source harnesses without verification, then workflow, and finally state support.
Supervision covers both newly targeted responsibilities and those addressed at earlier stages, with replay and cue dropout supplementing the curriculum.
We evaluate 9B and 35B models on SWE-QA and SWE-QA-Pro \citep{peng2026sweqa,cai2026sweqapro}, testing every training checkpoint under all four deployment harnesses.
Selected intermediate checkpoints achieve tools-only scores close to those of their respective starting checkpoints deployed with the full harness.
For the 9B model, the $H_2$-targeted checkpoint improves tools-only scores over the $H_4$-trained starting checkpoint by $5.01$ points on SWE-QA and $3.09$ points on SWE-QA-Pro.
The gains vary with model scale and deployment configuration, and further annealing does not uniformly improve performance.

Our contributions are threefold:
\begin{enumerate}[leftmargin=2em]
 \vspace{-0.2cm}
    \item We formulate \textbf{Harness Internalization} as a learning objective: transferring specified process-control responsibilities to the model while retaining task performance after the corresponding external support is withdrawn.
    We assess progress through task performance and control-related behavior under predefined support removals.
    \item We introduce \textbf{\method{} (HAT)}, which combines explicit control supervision with a curriculum over progressively weaker harness support, supplemented by replay and cue dropout, to train agents for deployment with less external guidance.

    \item We evaluate the \textbf{full checkpoint-by-harness matrix} at two model scales on SWE-QA and SWE-QA-Pro, showing improvements under weak support and revealing how preferred checkpoints vary with model scale and deployment objective.
\end{enumerate}
\section{Related Work}
\label{sec:related}
\vspace{-0.2cm}
\paragraph{Adapting the external harness.}
ADAS and Meta-Harness optimize agent programs, workflows, or
runtime code \citep{hu2025adas,lee2026metaharness}. HarnessX extends this
direction to composable processors and harness--model co-evolution,
while reporting evaluation-shortcut exploitation during optimization
\citep{chen2026harnessx}. These approaches retain the optimized runtime
at deployment. We instead use HarnessX's compositional substrate to
define a fixed removal ladder and transfer selected control
responsibilities to the policy.
\vspace{-0.2cm}
\paragraph{Learning from interaction trajectories.}
FireAct and AgentTuning train agents on interaction trajectories
\citep{chen2023fireact,zeng2023agenttuning}, while Agent-FLAN separates
reasoning from format learning and ATLaS prioritizes critical decision
steps \citep{chen2024agentflan,chen2025atlas}. We focus on who supplies
control decisions and which tokens receive supervision: harness
guidance provided only as input is not directly supervised as an
output. Our observation--reasoning--action records turn targeted control
responsibilities into model outputs while retaining observed evidence
as input. The Direct--Layered comparison evaluates this construction
jointly, not control supervision alone, because segmentation also
changes token weighting and optimizer exposure.
\vspace{-0.2cm}
\paragraph{Distilling guidance and learning harness policies.}
Knowledge and policy distillation provide a teacher--student perspective
\citep{hinton2015distillingknowledgeneuralnetwork,rusu2016policydistillation}.
Prompt distillation, Skill-SD, and workflow compilation transfer
document-, skill-, or workflow-based guidance into policies that operate
without that guidance at inference
\citep{kujanpaa2024promptdistillation,wang2026skillsd,dennis2026compilingworkflows}.
In contrast, Harness-1 retains environment-side search state
\citep{jiang2026harness1}, and EvoHarness-RL learns selective access to
Belief, Progress, and Experience, describing declining harness use as
annealing \citep{ning2026evoharness}. We combine explicit control-decision
targets with predetermined layer removals and evaluate every checkpoint
under every deployment harness, distinguishing reduced scaffold use
from retained capability after its removal.

\vspace{-0.2cm}

\section{Method}
\label{sec:method}
\vspace{-0.2cm}
\subsection{Problem Formalization}
\label{sec:formalization}
\vspace{-0.2cm}
\subsubsection{Execution of Agentic Tasks}
\label{sec:trace}
\vspace{-0.2cm}
Given an environment $E$ whose state is partially observable to the model, and an instruction $q$, a language model $\pi_\theta$ must interact with $E$ over multiple steps through a tool set $Tools=\{T_1,\dots,T_n\}$ to produce an outcome $\hat{y}$ (a predicted answer, patch or report). Each task can be regarded as a trace:
\begin{equation}
\tau=\big(s_0,\;a_1,o_1,\;a_2,o_2,\;\dots,\;a_i,o_i,\;\hat{y}\big),
\qquad
a_t\sim\pi_\theta\big(\cdot\mid q,\;h_{t-1}\big),
\quad
o_t=E(a_t),
\end{equation}
where $s_0$ is the initial environment state, $a_t$ is usually an action in $Tools$, $o_t$ is the observation returned by $E$, and $h_{t-1}=(a_1,o_1,\dots,a_{t-1},o_{t-1})$ is the interaction history. Compared with single-turn generation, agentic tasks exhibit three structural properties that make them hard:

\begin{enumerate}[leftmargin=1.6em,itemsep=2pt]
\vspace{-0.2cm}
  \item \textbf{Long horizon and partial observability.} The information required for solution is absent from the initial instruction and must be actively acquired by $\pi_{\theta}$ through a sequence of actions, each of which reveals only a local view of $E$.
  \item \textbf{Sparse credit assignment.} Intermediate actions carry no labels; only the final outcome is evaluated. $\pi_{\theta}$ must therefore judge on its own whether the evidence gathered so far is sufficient to commit to an answer.
\item \textbf{Groundedness.} The predicted outcome $\hat{y}$ should be supported by relevant observed evidence. Final-answer correctness alone does not establish whether the agent gathered or used that evidence.
\end{enumerate}

\vspace{-0.2cm}
For a given question and reference answer, let $J(\tau)$ denote the score assigned to the final answer $\hat{y}$ in trajectory $\tau$. We implement this score using a reference-based judge that receives the question, reference answer, and final answer (Appendix~\ref{sec:llm_judge_details}). The judge does not receive intermediate observations and therefore does not directly assess whether $\hat{y}$ is grounded in the evidence gathered during the trajectory. We separately examine selected control behaviors using the diagnostics in Appendix~\ref{app:decision-probe}.

\subsubsection{Formalizing Harness Operations}
\label{sec:harness}
We define a harness as a set of external processors $H=\{P_1,\dots,P_m\}$ usually acting on the model's context. At each step, the processors inject external content or state:
\begin{equation}
c_t \;=\; \mathcal{P}_H\big(c_{t-1},a_{t-1},o_{t-1}\big),
\qquad
a_t \sim \pi_\theta\big(\cdot \mid q,\;h_{t-1},\;c_t\big).
\end{equation}

The harness maintains an \emph{external state} $c_t$ and supplies it to the policy at runtime. We study four \emph{nested} harness configurations $H_1 \subset H_2 \subset H_3 \subset H_4$ (Table~\ref{tab:levels}). Each $H_\ell$ denotes the complete configuration; for $\ell\geq2$, $\Delta_\ell$ denotes its added control layer. Thus $H_3$ includes tools, state, and workflow, whereas $\Delta_3$ is workflow control.

\begin{table}[t]
\centering
\caption{Deployment hierarchy and targeted control responsibilities.
Each level retains all preceding support; bold marks the added component.
Tools, execution guards, and token budgets remain external throughout.
The verifier has no access to the reference answer.}
\label{tab:levels}
\vspace{-0.2cm}

\begingroup
\small
\definecolor{harnesshead}{HTML}{E8EFF5}
\definecolor{harnessstripe}{HTML}{F5F8FA}
\definecolor{harnessrule}{HTML}{AABAC7}

\setlength{\tabcolsep}{5pt}
\setlength{\extrarowheight}{4pt}
\setlength{\aboverulesep}{0pt}
\setlength{\belowrulesep}{0pt}
\renewcommand{\arraystretch}{1.18}
\renewcommand{\tabularxcolumn}[1]{m{#1}}
\arrayrulecolor{harnessrule}

\begin{tabularx}{\linewidth}{
  >{\raggedright\arraybackslash}m{0.25\linewidth}|
  >{\raggedright\arraybackslash}m{0.4\linewidth}
  >{\raggedright\arraybackslash}X
}
\toprule[1pt]
\rowcolor{harnesshead}
\textbf{Nested configuration}
& \textbf{Added runtime support}
& \textbf{Targeted responsibility} \\
\midrule[0.4pt]

$H_1=\text{\textbf{Tools}}$
& Tool access and execution safeguards
& Gather task evidence using external tools. \\[3pt]

\rowcolor{harnessstripe}
$H_2=H_1+\text{\textbf{State}}$
& Recent history, progress ledger, and repetition detection
& Track evidence and gaps; avoid unproductive repetition. \\[3pt]

$H_3=H_2+\text{\textbf{Workflow}}$
& Locate $\to$ Read $\to$ Connect $\to$ Answer
& Organize investigation and answer construction; choose phase transitions. \\[3pt]

\rowcolor{harnessstripe}
$H_4=H_3+\text{\textbf{Verification}}$
& Heuristic draft checks and bounded repair
& Check draft support; revise, gather evidence, or finish. \\[8pt]

\bottomrule[1pt]
\end{tabularx}

\arrayrulecolor{black}
\endgroup
\vspace{-0.3cm}
\end{table}

This layering maps onto the above three difficulties: $H_2$ mitigates state forgetting over long horizons, $H_3$ mitigates disorganized exploration under sparse credit assignment, and $H_4$ directly targets groundedness by gating the terminal step. Because the levels are nested, a higher level $H_i$ trace simultaneously contains the behaviors of lower levels $H_{\le i}$, providing the foundation for the layer-wise annealing introduced later.



\subsubsection{Harness Internalization Objective}
\label{sec:hgap}
For a policy $\pi_\theta$ and a deployment harness $H$, let $S(\pi_\theta,H)$ denote the score on a held-out evaluation set. We define the \textbf{harness-dependency gap} between two configurations as
\begin{equation}
HGap(\pi_\theta,i,j)
=
S(\pi_\theta,H_i)-S(\pi_\theta,H_j),
\qquad i>j.
\label{eq:hgap}
\end{equation}
A positive gap indicates that the additional support in $H_i$ improves performance; a negative gap indicates that it reduces performance.

We also measure variation across the full deployment ladder using
\begin{equation}
\Delta(\pi_\theta)
=
\max_{j\in\{1,2,3,4\}} S(\pi_\theta,H_j)
-
\min_{j\in\{1,2,3,4\}} S(\pi_\theta,H_j).
\label{eq:objective}
\end{equation}
Our goal is to achieve strong performance under weak support while reducing sensitivity to the deployment harness. We therefore assess $S(\pi_\theta,H_1)$ together with $\Delta(\pi_\theta)$ and the full score matrix. A small range alone is insufficient, because a policy that performs poorly under every configuration would also have low cross-harness variation. The training procedure below pursues this goal through supervised fine-tuning.

\begin{figure*}[t]
  \centering
  \includegraphics[width=\textwidth]{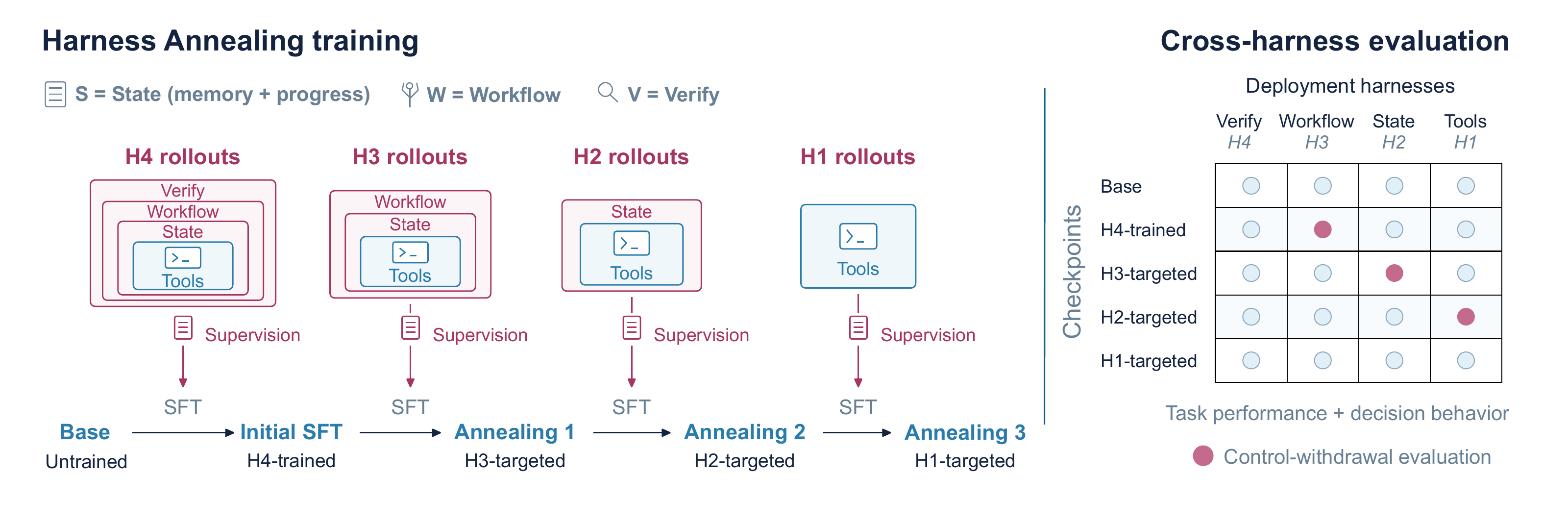}
  \vspace{-1cm}
  \caption{\textbf{Training and evaluation of HAT.}
  Initial SFT produces the Layered $H_4$-trained checkpoint. Three annealing
  stages then introduce records derived from $H_3$, $H_2$, and $H_1$ rollouts,
  supplemented by stronger-source replay and cue-dropout variants.
  The rollout labels identify source harnesses; targeted checkpoint labels
  identify cumulative training through the corresponding source level.
  Every checkpoint is evaluated under all four deployment harnesses.
  Red cells test removal of the control layer targeted at that stage:
  verification at $H_3$, workflow at $H_2$, and state at $H_1$.
  Related cue-removal patterns may occur in training through dropout.}
  \label{fig:method}
  \vspace{-0.2cm}
\end{figure*}

\subsection{Harness Annealing Training}
A natural baseline is supervised fine-tuning on complete tool-use trajectories. When harness decisions appear only as conditioning context, this objective does not explicitly train the policy to generate them. We introduce two components to address this gap: \emph{Layer-wise Reasoning Supervision}, which makes selected control decisions explicit targets, and \emph{Harness-Dependency Annealing}, which progressively introduces demonstrations collected under weaker harnesses into continued training.

\subsubsection{Layer-wise Reasoning Supervision}
\label{sec:supervision}
We construct supervision targets for control decisions at each harness level and use them to train the policy to act with weaker external support.

\vspace{-0.2cm}
\paragraph{Trajectory collection.}
For training tasks $\mathcal{D}_{\mathrm{train}}=\{(q_n,E_n,y_n)\}_{n=1}^{N}$,
we collect teacher trajectories under each source harness:
\begin{equation}
\mathcal{B}=\bigcup_{\ell=1}^{4}\mathcal{B}_\ell,
\qquad
\mathcal{B}_\ell=\big\{\tau_{n,r}^{(H_\ell,\pi_{\mathrm{tea}})}\big\}_{n,r},
\end{equation}
where $r$ indexes rollout rounds. The source bank $\mathcal{B}_\ell$ contains
raw trajectories collected under $H_\ell$; $\mathcal{D}_\ell$ denotes the
training records derived from them. The source index is preserved after
rewriting or cue dropout. Trajectories for the same task across harnesses
support comparisons of recorded decisions and interventions, although their
differences also reflect subsequent action choices and sampling variability.
Control annotations are constructed deterministically from recorded
interactions and level-specific rules, with no additional model calls.

\vspace{-0.2cm}
\paragraph{Triplet rewriting.}
At model decision boundaries, we organize the trajectory into
\textsc{Obs}, \textsc{Think}, and \textsc{Act} components:
\begin{itemize}[leftmargin=1.6em,itemsep=2pt]
  \item \textbf{\textsc{Obs}} contains the task, interaction history,
  observed evidence, and harness cues retained by the record construction.
  Its available support is specified separately from the source index.
  \item \textbf{\textsc{Think}} contains control-oriented supervision,
  such as workflow organization or answer checking, that the model is
  trained to produce. Related cues may also remain in the input.
  \item \textbf{\textsc{Act}} contains the recorded teacher action or answer.
\end{itemize}
For a record $x\in\mathcal{D}_\ell$, let $\mathcal{J}(x)$ denote its supervised
decision boundaries. The loss covers \textsc{Think} and \textsc{Act} tokens,
with the system prompt and designated special-token block masked out:
\begin{equation}
\mathcal{L}_{\mathrm{LSFT}}^{(\ell)}(\theta)
=-\mathbb{E}_{x\sim\mathcal{D}_\ell}
\sum_{t\in\mathcal{J}(x)}
\log\pi_\theta\big(\mathrm{Think}_t,\mathrm{Act}_t\mid\mathrm{Obs}_t\big).
\label{eq:lsft}
\end{equation}
\vspace{-0.2cm}
\subsection{Harness-Dependency Annealing}
\label{sec:annealing}

\paragraph{Checkpoint notation.}
An $H_\ell$-trained checkpoint, $\theta_\ell^{\mathrm{ind}}$, is trained
independently from Base using only $H_\ell$-source data; Direct and Layered
specify the supervision construction. An $H_\ell$-targeted checkpoint,
$\theta_\ell^{\mathrm{ann}}$, continues the cumulative curriculum through
source level $\ell$. Its index identifies the latest source level introduced,
not the deployment harness. Training starts from Layered $H_4$-trained,
so $\theta_4^{\mathrm{ann}}=\theta_4^{\mathrm{ind,Layered}}$.

\paragraph{Training curriculum.}
The three annealing stages introduce $\mathcal{D}_3$, $\mathcal{D}_2$, and
$\mathcal{D}_1$ in sequence. At stage $\ell$, the mixture $\mathcal{M}_\ell$
combines current-source records, stronger-source replay, and cue-dropout
variants where applicable (Appendix~\ref{app:training-mixtures}).
Let $\mathcal{L}_{\mathrm{mix}}^{(\ell)}$ denote the loss in
Equation~\ref{eq:lsft} evaluated over this mixture. Training proceeds as
\begin{equation}
\theta_0
\xrightarrow{\mathcal{L}_{\mathrm{LSFT}}^{(4)}}\theta_4^{\mathrm{ann}}
\xrightarrow{\mathcal{L}_{\mathrm{mix}}^{(3)}}\theta_3^{\mathrm{ann}}
\xrightarrow{\mathcal{L}_{\mathrm{mix}}^{(2)}}\theta_2^{\mathrm{ann}}
\xrightarrow{\mathcal{L}_{\mathrm{mix}}^{(1)}}\theta_1^{\mathrm{ann}}.
\label{eq:annealing-path}
\end{equation}
For example, $H_3$-targeted continues from $H_4$-trained using $H_3$-source
records and $H_4$-source replay. Standard $H_3$ records retain workflow support;
cue dropout reduces selected input cues. The objective is to learn workflow
control while retaining the verification behavior targeted earlier.
We test this transfer under $H_2$, where neither workflow nor verification
control is supplied externally.

The corresponding withdrawal evaluations are $H_4$-trained under $H_3$,
$H_3$-targeted under $H_2$, and $H_2$-targeted under $H_1$.
$H_1$-targeted continues training with tools-only source data; it does not
target removal of the tools themselves. Every checkpoint is evaluated under
all four harnesses. Appendix~\ref{app:training-paths} summarizes the paths
and distinguishes learning objectives from demonstrated internalization.

\vspace{-0.2cm}
\subsection{Design Rationale}
\label{sec:rationale}
\vspace{-0.2cm}

\paragraph{Input support and supervised control.}
\textsc{Obs} supplies the conditions available at a decision boundary,
whereas \textsc{Think} supervises control judgments. Making a judgment an
output target does not require removing every related input cue from every
record. Standard examples provide supported demonstrations; cue dropout
varies selected cues, and deployment evaluations test behavior after the
corresponding runtime components are removed. Tools and observed evidence
remain external throughout.

\vspace{-0.4cm}
\paragraph{Direction.}
The stronger-to-weaker curriculum introduces source data with progressively
less external control. It targets verification, then workflow, then state
control, while replay revisits earlier supervision. This ordering lets each
stage build on its predecessor rather than requiring the initial model to
learn all control responsibilities from tools-only demonstrations.
The final stage introduces tools-only source data, aligning its standard
inputs with the weakest deployment interface. Replay still retains stronger
support, and recorded histories may differ from those visited at deployment;
the curriculum therefore does not assume that source records and deployment
contexts are identical.

\section{Experiments}
\label{sec:experiments}
\vspace{-0.2cm}
In this section, experiments are conducted to answer the following research questions:

\begin{itemize}[leftmargin=2em]
    \vspace{-0.2cm}
    \item RQ1: Can Harness Annealing internalize external control, while retaining performance under weak support after harness layers are removed?
    \item RQ2: Does Layered supervision transfer better under weak support than trajectory-level SFT?
    \item RQ3: Does Harness Annealing improve performance on out-of-distribution tasks?
\end{itemize}

\vspace{-0.2cm}
\subsection{Experimental Setup}
\vspace{-0.4cm}

\noindent
\begin{minipage}[t]{0.35\linewidth}
\vspace{0pt}
\small
\textbf{Dataset.}
We conduct our experiments on repository-level question answering (repo-level QA), where an agent must answer questions by understanding an entire code repository rather than a single file or function. Agents may search and read the repository but may not modify it. We use the training split of SWE-QA-Pro~\citep{cai2026sweqapro} for training and evaluate on its test split and SWE-QA~\citep{peng2026sweqa}.
\end{minipage}
\hfill
\begin{minipage}[t]{0.63\linewidth}
\vspace{0pt}
\small
\refstepcounter{table}\label{tab:data-accounting}
\textbf{Table \thetable: Datasets and Rollout Configurations.} \newline Evaluation yields 2,880 episodes per checkpoint on SWE-QA and 1,040 on SWE-QA-Pro.
\par\smallskip
\centering
\setlength{\tabcolsep}{2.5pt}
\renewcommand{\arraystretch}{1.2}
\begin{tabular}{@{}
  >{\raggedright\arraybackslash}m{0.17\linewidth}
  >{\raggedright\arraybackslash}m{0.3\linewidth}
  >{\raggedleft\arraybackslash}m{0.15\linewidth}
  >{\raggedright\arraybackslash}
    m{\dimexpr0.36\linewidth-6\tabcolsep\relax}
@{}}
\toprule
Role & Dataset / split & Questions & Setup \\
\midrule
Training
& SWE-QA-Pro train
& 464
& 4 harness levels\newline $\times$ 3 rollout rounds \\[3pt]
Evaluation
& SWE-QA
& 720
& 4 harness levels \\[3pt]
Evaluation
& SWE-QA-Pro test
& 260
& 4 harness levels \\
\bottomrule
\end{tabular}
\end{minipage}
\par\medskip

\vspace{-0.4cm}
\paragraph{Harness Details}

We implement the 4 levels of harness with HarnessX~\citep{chen2026harnessx}, which represents prompts, memory, tools, and runtime control as composable processors \citep{harnessx2026report,harnessx2026software}.  We defer the full configuration details of HarnessX to Appendix~\ref{app:harness-details}.


\vspace{-0.2cm}
\paragraph{Implementation Details}
We fine-tune Qwen3.5-9B and Qwen3.5-35B-A3B~\citep{qwen2026qwen35modelcard}
on GPT-5.5 trajectories~\citep{openai2026gpt55} collected under all four
source harnesses. Base denotes the model before task-specific fine-tuning.
Annealing starts from Layered $H_4$-trained, obtained by fine-tuning Base
on records derived from $\mathcal{B}_4$ and also reported in
Table~\ref{tab:direct-layered}. Each subsequent stage initializes from
its predecessor and uses teacher-generated mixtures, without new
student rollouts. Training examples follow Section~\ref{sec:method},
with reasoning tokens supervised unless otherwise stated.
After quality filtering and removal of broken or duplicate records,
5,007 trajectories remain across the four source harnesses.
Stage mixtures use current-source records, with stronger-source replay
and cue-dropout variants where applicable.
Table~\ref{tab:mixtures} reports mixture weights based on the original
token proxy for 9B and tokenizer-counted supervised tokens for 35B.

Using LLaMA-Factory~\citep{zheng2024llamafactory}, each annealing stage
runs for one epoch with bf16, rank-64 LoRA, and a per-device batch size
of 1. The 65,536-token training cutoff exceeds the longest examples
(approximately 58K--61K tokens), so none are truncated.
Appendix~\ref{app:implementation-details} provides annealing configurations
and the 35B checkpoint-selection procedure.

SGLang~\citep{zheng2024sglang} serves rollout collection and evaluation.
Both evaluation modes use a 65,536-token context window.
No-thinking evaluation permits 4,096 completion tokens per call and
24 agent steps per episode; thinking-enabled evaluation permits
8,192 tokens and 32 steps. Results are not pooled across modes;
main annealing results use thinking-enabled evaluation unless otherwise
noted. Evaluation follows the reference-based SWE-QA rubric~\citep{peng2026sweqa},
summing five GPT-5.5 scores, each ranging from 1 to 20.

\begin{table*}[t]
\setlength{\abovecaptionskip}{0pt} 
\caption{\textbf{Every training checkpoint is evaluated under every deployment configuration.} Scores are mean reference-based judge scores; higher is better. Red cells test the control responsibility targeted at each stage after its external support is removed: $H_4$-trained under $H_3$, $H_3$-targeted under $H_2$, and $H_2$-targeted under $H_1$. Related cue-removal patterns may occur in training through dropout. $H_1$ exposes the bare tool interface and each subsequent harness adds one layer of control, so that $H_1 \subset H_2 \subset H_3 \subset H_4$. Within each model--benchmark block, bold marks the best in each column. The starting row is Layered $H_4$-trained; each targeted checkpoint continues from the row above.}
\label{tab:main-results}
\begin{center}
\footnotesize
\setlength{\tabcolsep}{4pt}
\renewcommand{\arraystretch}{1.08}
\begin{tabular*}{\textwidth}{@{\extracolsep{\fill}}lllrrrrrrr@{}}
\toprule
\multirow{2}{*}{Model} & \multirow{2}{*}{Benchmark} & \multirow{2}{*}{Checkpoint}
 & \multicolumn{4}{c}{Deployment Harness} & \multicolumn{3}{c}{Row summary} \\
\cmidrule(lr){4-7} \cmidrule(l){8-10}
 & &                   & $H_1$                  & $H_2$                 & $H_3$                 & $H_4$             & Mean              & Worst             & $\Delta$ \\
\midrule
\multirow{10}{*}{Qwen3.5-9B}
 & \multirow{5}{*}{SWE-QA}
 &   Base              & 59.93                  & 62.92                 & 67.28                 & 66.79             & 64.23             & 59.93             & 7.35             \\
 & & $H_4$-trained     & 71.99                  & 75.94                 & \frontiercell{\textbf{77.47}}  & 75.70             & 75.28             & 71.99             & 5.48             \\
 & & $H_3$-targeted    & \textbf{77.41}         & \frontiercell{\textbf{77.15}}  & 75.10                 & 75.66             & 76.33             & 75.10             & 2.31             \\
 & & $H_2$-targeted    & \frontiercell{77.00}   & 76.62                 & 76.43                 & \textbf{76.45}    & \textbf{76.63}    & \textbf{76.43}    & \textbf{0.57}    \\
 & & $H_1$-targeted    & 76.30                  & 75.12                 & 74.43                 & 74.16             & 75.00             & 74.16             & 2.14             \\
\addlinespace[2pt]
\cmidrule(l){2-10}
 & \multirow{5}{*}{SWE-QA-Pro}
 &   Base              & 57.96                  & 63.19                 & 66.40                 & 66.39             & 63.49             & 57.96             & 8.44             \\
 & & $H_4$-trained     & 74.64                  & 77.81                 & \frontiercell{76.87}  & 78.25             & 76.89             & 74.64             & 3.61             \\
 & & $H_3$-targeted    & 77.06                  & \frontiercell{\textbf{77.96}}  & 75.97                 & 77.54             & 77.13             & 75.97             & 1.99             \\
 & & $H_2$-targeted    & \frontiercell{\textbf{77.73}}   & 77.83                 & 76.98                 & \textbf{78.30}    & \textbf{77.71}    & 76.98             & 1.32             \\
 & & $H_1$-targeted    & 77.12                  & 77.40                 & \textbf{77.94}        & 77.47             & 77.48             & \textbf{77.12}    & \textbf{0.82}    \\
 \cmidrule(l){1-10}
\multirow{10}{*}{Qwen3.5-35B}
 & \multirow{5}{*}{SWE-QA}
 &   Base              & 78.14                  & 75.66                 & 75.72                 & 76.23             & 76.44             & 75.66             & 2.48             \\
 & & $H_4$-trained     & 80.60         & 79.85                 & \frontiercell{80.13}  & \textbf{81.15}    & 80.43    & 79.85    & \textbf{1.30}    \\
 & & $H_3$-targeted    & 80.33                  & \frontiercell{\textbf{81.71}}  & 79.61                 & 79.71             & 80.34             & 79.61             & 2.10             \\
 & & $H_2$-targeted    & \frontiercell{\textbf{81.00}} & 81.48 & 80.04 & 80.42 & \textbf{80.74} & \textbf{80.04} & 1.44 \\
 & & $H_1$-targeted    & 80.51 & 81.23 & 79.62 & 81.03 & 80.60 & 79.62 & 1.61 \\
 \addlinespace[2pt]
\cmidrule(l){2-10}
 & \multirow{5}{*}{SWE-QA-Pro}
 &   Base              & 72.16                     & 73.58                    & 74.30                    & 73.53                & 73.39                & 72.16                & 2.14               \\
 & & $H_4$-trained     & 81.82                     & 82.72                    & \frontiercell{\textbf{82.48}}                    & \textbf{83.95}                & \textbf{82.74}                & \textbf{81.82}                & 2.13               \\
 & & $H_3$-targeted    & 79.88                     & \frontiercell{\textbf{82.96}}                    & 81.61                    & 82.03                & 81.62                & 79.88                & 3.08               \\
 & & $H_2$-targeted    & \frontiercell{\textbf{83.18}}                     & 81.65                    & 81.52                    & 81.85                & 82.05                & 81.52                & 1.66               \\
 & & $H_1$-targeted    & 81.06                     & 82.42                    & 82.20                    & 81.77                & 81.86                & 81.06                & \textbf{1.36}               \\
\bottomrule
\end{tabular*}
\end{center}
\vspace{-0.5cm}
\end{table*}


\subsection{Main Results}
\vspace{-0.2cm}
Table~\ref{tab:main-results} compares every checkpoint under all four deployment harnesses.
\vspace{-0.4cm}
\paragraph{Annealing improves performance under weak support.}
For 9B, relative to the $H_4$-trained starting checkpoint, $H_2$-targeted improves $H_1$ scores from $71.99$ to $77.00$ on SWE-QA and from $74.64$ to $77.73$ on SWE-QA-Pro, gains of $5.01$ and $3.09$ points. Its cross-harness range $\Delta$ decreases from $5.48$/$3.61$ to $0.57$/$1.32$ on the two benchmarks. Thus, reduced variation accompanies stronger performance under weak support. For 35B, $H_1$ scores improve by $0.40$ and $1.36$ points, but $\Delta$ changes from $1.30$ to $1.44$ on SWE-QA and from $2.13$ to $1.66$ on SWE-QA-Pro. Better weak-support performance therefore does not always imply lower cross-harness variation.
\begin{table*}[t]
\setlength{\abovecaptionskip}{0pt} 
\caption{\textbf{Direct and Layered supervision produce different cross-harness transfer profiles.}
Scores are mean reference-based judge scores; higher is better. For each model scale, the table reports one Base row, four Direct-trained checkpoints, and four independently Layered-trained checkpoints. All adapted checkpoints are trained independently from Base using the indicated source level. Columns specify the deployment harness. Bold marks the highest score among the Layered checkpoints in each deployment column for each model size.}
\label{tab:direct-layered}
\begin{center}
\scriptsize
\setlength{\tabcolsep}{4.8pt}
\renewcommand{\arraystretch}{1.05}
\resizebox{\textwidth}{!}{%
\begin{tabular}{@{}lllrrrrrrrr@{}}
\toprule
\multirow{2}{*}{Model} & \multirow{2}{*}{Supervision} & \multirow{2}{*}{Independent checkpoint} & \multicolumn{4}{c}{SWE-QA inference harness} & \multicolumn{4}{c}{SWE-QA-Pro inference harness} \\
\cmidrule(lr){4-7}\cmidrule(lr){8-11}
 & & & $H_1$ & $H_2$ & $H_3$ & $H_4$ & $H_1$ & $H_2$ & $H_3$ & $H_4$ \\
\midrule
\multirow{9}{*}{9B}
 & \multirow{5}{*}{Direct}
 & Base              & 59.93 & 62.92 & 67.28 & 66.79 & 57.96 & 63.19 & 66.40 & 66.39 \\
 & & $H_1$-trained   & 32.45 & 65.83 & 62.29 & 59.65 & 34.74 & 65.69 & 66.28 & 64.02 \\
 & & $H_2$-trained   & 16.76 & 57.83 & 58.98 & 61.30 & 17.88 & 66.95 & 62.58 & 62.63 \\
 & & $H_3$-trained   & 17.91 & 64.15 & 64.60 & 61.71 & 18.43 & 66.89 & 66.95 & 66.88 \\
 & & $H_4$-trained   & 16.51 & 60.21 & 61.28 & 62.19 & 16.69 & 68.88 & 67.23 & 64.02 \\
\cmidrule(l){2-11}
 & \multirow{4}{*}{Layered}
 & $H_1$-trained   & \textbf{72.48} & 75.18 & 75.78 & 72.91 & \textbf{77.52} & \textbf{78.28} & \textbf{78.02} & 76.72 \\
 & & $H_2$-trained   & 53.73 & 69.23 & 68.56 & 66.18 & 59.75 & 71.53 & 68.43 & 66.21 \\
 & & $H_3$-trained   & 62.27 & 67.52 & 72.08 & 55.14 & 68.70 & 73.21 & 68.97 & 61.54 \\
 & & $H_4$-trained   & 71.99 & \textbf{75.94} & \textbf{77.47} & \textbf{75.70} & 74.64 & 77.81 & 76.87 & \textbf{78.25} \\
\midrule
\multirow{9}{*}{35B}
 & \multirow{5}{*}{Direct}
   & Base            & 78.14 & 75.66 & 75.72 & 76.23 & 72.16 & 73.58 & 74.30 & 73.53 \\
 & & $H_1$-trained   & 30.62 & 63.35 & 62.27 & 60.12 & 31.57 & 65.63 & 66.10 & 60.74 \\
 & & $H_2$-trained   & 21.14 & 68.32 & 68.74 & 65.09 & 23.48 & 75.14 & 73.85 & 74.48 \\
 & & $H_3$-trained   & 25.70 & 70.39 & 69.18 & 67.42 & 32.31 & 75.51 & 77.59 & 74.72 \\
 & & $H_4$-trained   & 24.58 & 69.93 & 71.16 & 66.57 & 30.83 & 76.16 & 76.58 & 74.23 \\
\cmidrule(l){2-11}
 & \multirow{4}{*}{Layered}
 & $H_1$-trained   & 80.51 & 75.49 & 76.27 & 74.91 & 81.01 & 79.04 & 77.88 & 78.68 \\
 & & $H_2$-trained   & 73.68 & 75.12 & 75.69 & 73.59 & 73.59 & 78.62 & 75.88 & 76.60 \\
 & & $H_3$-trained   & 65.81 & 77.88 & 76.43 & 76.48 & 71.80 & 80.88 & 77.73 & 78.43 \\
 & & $H_4$-trained   & \textbf{80.60} & \textbf{79.85} & \textbf{80.13} & \textbf{81.15} & \textbf{81.82} & \textbf{82.72} & \textbf{82.48} & \textbf{83.95} \\
\bottomrule
\end{tabular}%
}
\end{center}
\vspace{-0.4cm}
\end{table*}
\vspace{-0.2cm}
\paragraph{Performance after removal of the targeted control layer.}
The red cells evaluate each checkpoint with the control layer targeted for internalization removed. For example, $H_3$-targeted is tested under $H_2$, with workflow and verification control both absent. For the 9B model, $H_4$-trained under $H_3$, $H_3$-targeted under $H_2$, and $H_2$-targeted under $H_1$ score $77.47$, $77.15$, and $77.00$ on SWE-QA and $76.87$, $77.96$, and $77.73$ on SWE-QA-Pro. Every red cell far exceeds Base under the \emph{same} weak deployment, with the gap widening as support weakens. These scores show strong task performance after the targeted support is removed. The red cells vary both checkpoint and deployment harness, so they characterize the withdrawal evaluations rather than isolate the gain from each training stage. The trajectory comparison in Appendix~\ref{app:trajectory-analysis} examines changes in model behavior after annealing.

\vspace{-0.3cm}
\paragraph{The preferred checkpoint depends on model scale and deployment objective.}
For the 9B model, $H_2$-targeted achieves the highest mean across the two benchmarks and four deployment harnesses, at $77.17$. For the 35B model, $H_4$-trained has the highest corresponding mean, at $81.59$, compared with $81.39$ for $H_2$-targeted. Nevertheless, $H_2$-targeted achieves the highest $H_1$ score on both benchmarks at 35B. Annealing therefore improves the weakest-support endpoint without necessarily maximizing performance averaged across all deployment configurations.

\vspace{-0.3cm}
\paragraph{Control-related behavior under cue removal.}
In the paired behavioral probe (Appendix~\ref{app:decision-probe}), stop/continue macro-F1 under $H_1$ input cues rises from $64.43$ at $H_4$-trained to $87.79$ at $H_1$-targeted. This teacher-forced probe holds reference states fixed and varies input cues, measuring action selection conditioned on reference reasoning rather than executing the runtime harnesses.

\subsection{Direct versus Layered Supervision}
\vspace{-0.2cm}





\textbf{Layered supervision avoids the weak-support collapse of Direct SFT.}
Table~\ref{tab:direct-layered} compares Direct and Layered supervision for each source-harness level. Direct supervision serializes complete trajectories, whereas Layered supervision constructs decision-level records as described in Section~\ref{sec:supervision}. All adapted checkpoints in this table are trained independently from Base using only the indicated source level; Direct and Layered specify the construction. The Layered $H_4$-trained row is the starting checkpoint in Table~\ref{tab:main-results}.

The two constructions diverge sharply under the weakest deployment $H_1$. For every training source, on both benchmarks and at both model scales, Layered supervision outperforms Direct supervision at $H_1$. The gap is large: for the 9B model on SWE-QA, the four Direct checkpoints span $16.51$--$32.45$ across sources, compared with $53.73$--$72.48$ for their Layered counterparts. Direct supervision thus collapses precisely where external support is weakest, whereas Layered supervision retains usable performance. The $H_1$-source Direct checkpoint also scores only $32.45$ under $H_1$, compared with $59.93$ for Base, showing that the degradation is not confined to a mismatch between source and deployment harnesses. This advantage over Direct is uniform at $H_1$; relative to Base, however, it is source-dependent, as the strongest source ($H_4$-trained) exceeds Base under every deployment while some weaker sources fall below Base under certain configurations.

\vspace{-0.2cm}
\subsection{Cross-Task OOD Generalization}
\label{sec:tau3_ood}
\vspace{-0.2cm}
\begin{figure*}[t]
  \centering
  \includegraphics[width=0.99\textwidth]{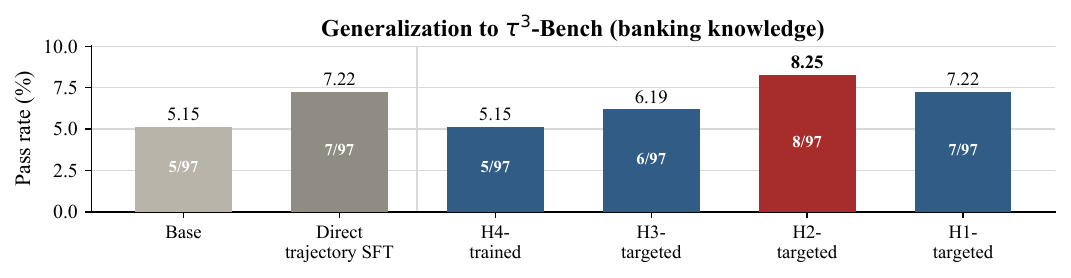}
    \vspace{-0.4cm}
    \caption{\textbf{Cross-task evaluation on the banking-knowledge domain of $\tau^3$-Bench.}
Zero-shot pass rates for Qwen3.5-9B on $97$ tasks under a fixed banking harness. Numbers inside bars indicate solved tasks.}
  \label{fig:ood}
  \vspace{-0.5cm}
\end{figure*}

We evaluate Qwen3.5-9B checkpoints zero-shot on the $97$ banking-knowledge tasks of $\tau^3$-Bench~\citep{barres2025tau2benchevaluatingconversationalagents,shi2026tauknowledgeevaluatingconversationalagents}. Fine-tuning uses only SWE-QA-Pro training trajectories. All checkpoints use the same banking harness, thinking mode, and $65{,}536$-token context window.

\vspace{-0.2cm}
Figure~\ref{fig:ood} suggests that harness annealing can improve transfer beyond repository QA without target-domain fine-tuning. The peak at an intermediate stage mirrors the in-domain trend, suggesting that further annealing does not necessarily improve transfer. However, the modest gain over Direct SFT and unmatched training budgets do not isolate control internalization as the source of this benefit.

\FloatBarrier

\section{Conclusion}
\vspace{-0.2cm}
We view model learning and harness development as a feedback loop: stronger models enable richer workflows, while harness-supported experience can train subsequent models. Yet learning from this experience does not automatically transfer the control decisions supplied by the harness. \method{} studies this objective by constructing explicit control supervision and progressively introducing demonstrations collected under weaker harnesses. On SWE-QA and SWE-QA-Pro, selected annealed checkpoints improve performance under weak support, although gains depend on model scale and deployment configuration and further annealing is not uniformly beneficial. These findings suggest a path for model--harness co-evolution: training can reshape the division of responsibility between the model and its harness. When a model can reliably assume a control responsibility, the corresponding guidance need no longer be repeatedly supplied at runtime. Harness development can then focus on the model's remaining limitations, enabling the model--harness system to tackle increasingly demanding tasks.

\bibliography{references}
\bibliographystyle{plainnat}

\clearpage
\appendix
\raggedbottom
\section{Limitations}

Our experiments use two Qwen3.5 models, one fixed harness hierarchy, and a single training seed. Evaluation focuses on repository question answering, with one banking-domain transfer experiment. This leaves open whether the same training recipe transfers to other model families, task domains, or harness designs.
Later checkpoints receive more training and use different data mixtures. Without equal-budget controls, the observed differences cannot be attributed to the annealing order alone, nor can the contributions of replay and cue dropout be separated. Direct and Layered supervision are also not matched for supervised-token counts or optimizer updates; their comparison therefore evaluates two training recipes rather than trajectory segmentation alone.

\section{Implementation Details}
\label{app:implementation-details}

The nested runtime configurations are $H_1$ (tools), $H_2$ (tools and
state), $H_3$ (tools, state, and workflow), and $H_4$ (all three plus
verification). The curriculum starts from Layered $H_4$-trained and produces
$H_3$-, $H_2$-, and $H_1$-targeted checkpoints. These labels identify
cumulative training history (Appendix~\ref{app:training-paths}).

\subsection{Training configuration}

\paragraph{Qwen3.5-9B.}
All three annealing stages continue the preceding LoRA adapter for one epoch on four
GPUs. One example per GPU and four accumulation steps give a global batch
size of 16. We use bf16, AdamW (betas 0.9/0.999, epsilon $10^{-8}$), a peak
learning rate of $10^{-6}$, cosine decay, a 0.05 warmup ratio, and seed 42.
The cutoff is 65,536 tokens. LoRA rank, alpha, and dropout are 64, 128,
and 0.05, with no bias and the attention/MLP targets recorded in the adapter
configurations. The three stages use 1,775, 2,049, and 2,291 records and
111, 129, and 144 optimizer updates. The recorded environment uses PEFT
0.19.1, Transformers 5.8.1, PyTorch 2.12.1+cu130, Datasets 4.8.4, and
Tokenizers 0.22.2.

\paragraph{Qwen3.5-35B-A3B.}
We use six GPUs with FSDP full sharding and gradient checkpointing.
One example per GPU and two accumulation steps give a global batch size
of 12. Each stage trains for one epoch. Peak learning rates are
$7.5\times10^{-7}$, $4.0\times10^{-7}$, and $2.5\times10^{-7}$ for
$H_3$-, $H_2$-, and $H_1$-targeted training, respectively.
We use bf16, fused AdamW, cosine decay, a 0.03 warmup ratio, weight decay
0.01, gradient clipping at 1.0, and training/data seed 42. LoRA rank,
alpha, and dropout are 64, 128, and 0.05. The configuration targets all
supported linear modules and freezes the vision tower and multimodal
projector. The \texttt{qwen3\_5} template preserves reasoning; prompts are
not supervised, assistant history is not masked, and packing is disabled.
The cutoff is 65,536 tokens. There are 1,812, 1,870, and 1,442 training
records and 151, 156, and 121 optimizer updates.

For 35B, 43 training-source question IDs are held out across levels;
checkpoint selection uses their validation loss, not benchmark test scores.
Validation runs approximately every quarter epoch. The selected updates are
151/151, 117/156, and 121/121: the $H_2$-targeted adapter is selected at
0.75 epoch, while the other two are selected at the final update.
Each subsequent stage loads the selected predecessor.

\subsection{Training mixtures}
\label{app:training-mixtures}

Table~\ref{tab:mixtures} separates current-source examples from stronger-source replay.
Every row sums to 100\%; a dash means that source is not used.
The 9B weights use the original token proxy, whereas 35B uses
tokenizer-counted supervised tokens.

\begin{table}[H]
\caption{\textbf{Current-source data and replay at each annealing stage (\%).}
Training proceeds down each block from $H_3$- to $H_1$-targeted.
The first stage starts from $H_4$-trained; subsequent stages continue the
preceding adapter. For $H_\ell$-targeted, the current source is $\mathcal{D}_\ell$; replay columns identify data sources, not deployment harnesses.}
\label{tab:mixtures}
\begin{center}
\appendixTableStyle
\begin{tabularx}{\linewidth}{lXrr!{\color{appendixRule}\vrule width0.35pt}rrr}
\appendixTopRule
\rowcolor{appendixHeader}
& & \multicolumn{2}{c!{\color{appendixRule}\vrule width0.35pt}}{\textbf{Current-source examples}} & \multicolumn{3}{c}{\textbf{Replay source}} \\
\rowcolor{appendixHeader}
\textbf{Model} & \textbf{Checkpoint} & Standard & Cue dropout & $\mathcal{D}_2$ & $\mathcal{D}_3$ & $\mathcal{D}_4$ \\
\midrule[0.45pt]
\textbf{9B} & $H_3$-targeted & 70 & 10 & -- & -- & 20 \\
   & $H_2$-targeted & 60 & 10 & -- & 20 & 10 \\
   & $H_1$-targeted & 50 & -- & 25 & 15 & 10 \\
\midrule[0.45pt]
\rowcolor{appendixWash}
\textbf{35B} & $H_3$-targeted & 40 & 30 & -- & -- & 30 \\
\rowcolor{appendixWash}
    & $H_2$-targeted & 20 & 30 & -- & 10 & 40 \\
\rowcolor{appendixWash}
    & $H_1$-targeted & 50 & -- & 20 & 10 & 20 \\
\appendixBottomRule
\end{tabularx}
\end{center}
\end{table}

Cue dropout modifies current-source inputs while retaining their response
labels. For 35B $H_3$ data, it removes workflow instructions and the
additional workflow observation to form an $H_2$-style input.
For $H_2$ data, it removes state instructions, the progress ledger, and
appended harness state to form an $H_1$-style input.
These variants retain their source index: $\mathcal{D}_3^{\mathrm{drop}}$ and $\mathcal{D}_2^{\mathrm{drop}}$ remain $H_3$- and $H_2$-source data, respectively. They preserve the recorded interaction history and are not new lower-harness rollouts.
Whole-record sampling makes realized 35B proportions slightly different
from the target percentages shown in the table.

\section{Training Paths and Control-Withdrawal Evaluation}
\label{app:training-paths}

Table~\ref{tab:training-paths} separates cumulative training history from
deployment. An $H_\ell$-trained checkpoint is trained independently from
Base on one source level. An $H_\ell$-targeted checkpoint continues the
curriculum from Layered $H_4$-trained through source level $\ell$, with
the mixtures in Appendix~\ref{app:training-mixtures}. Thus $H_3$-targeted
has the sequential history $H_4\to H_3$, whereas $H_3$-trained has only
$H_3$-source training. The index identifies the latest source introduced,
not the deployment harness.

\begin{table}[H]
\caption{\textbf{Cumulative training paths and corresponding control-withdrawal evaluations.}
Arrows indicate sequential introduction of source data through continued
training. Table~\ref{tab:mixtures} gives the mixtures used at each stage.}
\label{tab:training-paths}
\centering
\appendixTableStyle
\setlength{\tabcolsep}{5pt}
\begin{tabularx}{\linewidth}{
  @{}>{\raggedright\arraybackslash}p{0.17\linewidth}
  >{\raggedright\arraybackslash}p{0.23\linewidth}
  >{\raggedright\arraybackslash}p{0.26\linewidth}
  >{\raggedright\arraybackslash}X@{}
}
\appendixTopRule
\rowcolor{appendixHeader}
\textbf{Checkpoint} & \textbf{Cumulative source path}
& \textbf{Learning objective} & \textbf{Withdrawal evaluation} \\
\midrule[0.45pt]
$H_4$-trained\newline Initial SFT
& $H_4$
& Learn verification control.
& $H_3$: verification removed; tools, state, and workflow retained. \\
\rowcolor{appendixWash}
$H_3$-targeted\newline Annealing stage 1
& $H_4\to H_3$
& Learn workflow control; retain earlier verification behavior.
& $H_2$: workflow and verification absent; tools and state retained. \\
$H_2$-targeted\newline Annealing stage 2
& $H_4\to H_3\to H_2$
& Learn state control; retain verification and workflow behavior.
& $H_1$: state, workflow, and verification absent; tools retained. \\
\rowcolor{appendixWash}
$H_1$-targeted\newline Annealing stage 3
& $H_4\to H_3\to H_2\to H_1$
& Consolidate autonomous execution using tools-only source data.
& $H_1$: evaluate the tools-only endpoint; no further layer is removed. \\
\appendixBottomRule
\end{tabularx}
\end{table}

Source labels survive rewriting and cue dropout. Standard $H_3$ records
retain workflow cues even though $H_3$-targeted is tested under $H_2$:
training inputs and deployment support are specified separately.
The table states learning objectives; checkpoint names do not establish
successful internalization. Every checkpoint is evaluated under all four
harnesses. $H_1$-targeted remains a tools-only endpoint, not a step toward
tool-free $H_0$ deployment.

\section{Harness Details}
\label{app:harness-details}

Table~\ref{tab:harness-details} shows which support is available in each
deployment. All four expose \texttt{Read}, \texttt{Grep}, \texttt{Glob}, and read-only
\texttt{Bash}; higher levels add state, workflow, and answer-checking support. The question-only $H_0$ setting has no
repository context or tools and is not part of the four-column evaluation.

\begin{table}[H]
\caption{\textbf{Runtime support retained at each harness level.}
A checkmark means the component is present; a dash means it is absent.
Each higher level retains all lower-level support.}
\label{tab:harness-details}
\begin{center}
\appendixTableStyle
\begin{tabularx}{\linewidth}{>{\raggedright\arraybackslash}p{0.52\linewidth}!{\color{appendixRule}\vrule width0.35pt}*{4}{>{\centering\arraybackslash}X}}
\appendixTopRule
\rowcolor{appendixHeader}
\textbf{Runtime support} & $\boldsymbol{H_1}$ & $\boldsymbol{H_2}$ & $\boldsymbol{H_3}$ & $\boldsymbol{H_4}$ \\
\midrule[0.45pt]
Repository tools & \apYes & \apYes & \apYes & \apYes \\
Tool guards and token budget & \apYes & \apYes & \apYes & \apYes \\
\rowcolor{appendixWash}
Recent history and progress ledger & \apNo & \apYes & \apYes & \apYes \\
\rowcolor{appendixWash}
Repeated-action detection & \apNo & \apYes & \apYes & \apYes \\
Locate $\rightarrow$ Read $\rightarrow$ Connect $\rightarrow$ Answer & \apNo & \apNo & \apYes & \apYes \\
\rowcolor{appendixWash}
Answer verification and repair & \apNo & \apNo & \apNo & \apYes \\
\appendixBottomRule
\end{tabularx}
\end{center}
\end{table}

Tool guards and token budgets remain active in all four configurations.
$H_2$ bundles recent-history retrieval, a progress ledger, and loop
detection; this bundle is evaluated as one layer.
$H_4$ checks candidate answers and permits at most two keyword-based repair
searches. It may reject short, uncertainty-marked, or ungrounded drafts,
but never receives the reference answer.

\section{Evaluation with the LLM Judge}
\label{sec:llm_judge_details}
We evaluate final-answer quality on SWE-QA and SWE-QA-Pro using a reference-based LLM-as-a-Judge protocol with \texttt{gpt-5.5} as the judge. For each example, the judge receives the question $q_i$, reference answer $r_i$, and candidate final answer $a_i$, without the agent's interaction history or tool observations:
\[
  (q_i, r_i, a_i),
\]

For each example, the judge assigns five equally weighted integer scores:
\[
  s_{i,\mathrm{corr}},\; s_{i,\mathrm{comp}},\; s_{i,\mathrm{rel}},\; s_{i,\mathrm{clar}},\; s_{i,\mathrm{reas}} \in \{1,\ldots,20\},
\]
corresponding to correctness, completeness, relevance, clarity, and reasoning, respectively. The dimensions are defined as follows.

\begin{table}[H]
\setlength{\abovecaptionskip}{0pt}
\appendixTableStyle
\caption{\textbf{The five questions asked by the answer judge.}
Each dimension contributes 1--20 points. The full rubric and score anchors
are reproduced in the prompt below.}
\label{tab:dimensions}
\begin{center}
\vspace{+0.2cm}
\begin{tabularx}{\linewidth}{>{\raggedright\arraybackslash}p{0.18\linewidth}>{\raggedright\arraybackslash}X>{\raggedright\arraybackslash}p{0.26\linewidth}}
\appendixTopRule
\rowcolor{appendixHeader}
\textbf{Dimension} & \textbf{What is checked?} & \textbf{What lowers the score?} \\
\midrule[0.45pt]
Correctness & Do the claims agree with the reference? & Errors or contradictions \\
\rowcolor{appendixWash}
Completeness & Are the reference's key points covered? & Missing mechanisms or conditions \\
Relevance & Does the answer address the question? & Off-topic or irrelevant content \\
\rowcolor{appendixWash}
Clarity & Is the answer precise and easy to follow? & Ambiguity or poor organization \\
Reasoning & Do the stated claims support the conclusion? & Logical gaps or unjustified inferences \\
\appendixBottomRule
\end{tabularx}
\end{center}
\end{table}

The total judge score for example \(i\) is the unweighted sum
\begin{equation}
  S_i = s_{i,\mathrm{corr}} + s_{i,\mathrm{comp}} + s_{i,\mathrm{rel}} + s_{i,\mathrm{clar}} + s_{i,\mathrm{reas}}, \qquad S_i \in [5,100].
\end{equation}
No additional weighting, normalization, or learned calibration is applied, and the primary judge does not emit a binary pass/fail decision.

\subsection{Complete Judge Prompt}

The following is the complete prompt template used by the principal \texttt{gpt-5.5} judge. The placeholders \texttt{\{question\}}, \texttt{\{reference\}}, and \texttt{\{candidate\}} are replaced by the corresponding example fields.

\begin{judgepromptbox}
You are a STRICT and RIGOROUS evaluator. You must rate the candidate answer STRICTLY against the reference answer. Be CONSERVATIVE with high scores - only award high scores (16-20) when the candidate answer is truly excellent and closely matches the reference answer in quality and content.

CRITICAL EVALUATION PRINCIPLES:
1. Compare the candidate answer DIRECTLY with the reference answer point by point
2. Any deviation, omission, or inaccuracy should result in score reduction
3. High scores (16-20) should be RARE - reserve them only for answers that are nearly perfect
4. Be strict about factual accuracy - even minor errors should lower the correctness score
5. Missing key points from the reference answer should significantly reduce completeness score
6. Vague or imprecise language should lower clarity scores
7. When in doubt between two score ranges, choose the LOWER one

Evaluation Criteria and Scoring Guidelines (each scored 1 to 20, total score 100):
        1. Correctness (STRICT - penalize any inaccuracies):
            20 — ONLY if completely correct with ALL core points and details accurate, matching reference answer precisely
            16-19 — Mostly correct but must have only TRIVIAL inaccuracies; any noticeable error reduces to 15 or below
            12-15 — Partially correct; has some errors or omissions that affect understanding; main points may be accurate but details are wrong
            8-11 — Several errors or ambiguities that significantly affect understanding of core information
            4-7 — Many errors; misleading or fails to convey key information correctly
            1-3 — Serious errors; completely wrong or misleading
        2. Completeness (STRICT - penalize missing information):
            20 — ONLY if covers ALL key points from reference answer without ANY omission; must match reference in depth
            16-19 — Covers most key points but missing some non-trivial information; minor omissions are acceptable
            12-15 — Missing several important key points; content is noticeably incomplete compared to reference
            8-11 — Important information largely missing; content is one-sided or superficial
            4-7 — Covers very little relevant information; seriously incomplete
            1-3 — Covers almost no relevant information; completely incomplete
        3. Relevance (STRICT - penalize off-topic content):
            20 — ONLY if content is fully focused on question topic with NO irrelevant information whatsoever
            16-19 — Mostly focused but may have minor peripheral information; any significant off-topic content reduces score
            12-15 — Generally on topic but contains some off-topic content that detracts from answer
            8-11 — Topic not sufficiently focused; contains considerable off-topic or tangential content
            4-7 — Content deviates from topic; includes excessive irrelevant information
            1-3 — Majority of content irrelevant to the question
\end{judgepromptbox}

\Needspace{10\baselineskip}
\begin{judgepromptbox}[Verbatim judge prompt (continued)]
        4. Clarity (STRICT - penalize unclear expression):
            20 — ONLY if language is exceptionally fluent, clear, and precise; very easy to understand without any ambiguity
            16-19 — Mostly fluent and clear but may have minor unclear points; any significant ambiguity reduces score
            12-15 — Generally clear but some expressions are unclear or not concise; may require effort to understand
            8-11 — Expression somewhat awkward; has ambiguity or lacks fluency that hinders understanding
            4-7 — Language obscure; sentences are not smooth; significantly hinders understanding
            1-3 — Expression confusing; very difficult to understand
        5. Reasoning (STRICT - penalize weak logic):
            20 — ONLY if reasoning is exceptionally clear, logical, and well-structured; argumentation is excellent and matches reference quality
            16-19 — Reasoning is clear and logical with solid argumentation; minor logical gaps may exist
            12-15 — Reasoning generally reasonable but has noticeable logical jumps or organization issues
            8-11 — Reasoning is average; has logical jumps or organization problems that affect understanding
            4-7 — Reasoning unclear; lacks logical order; difficult to follow
            1-3 — No clear reasoning; logic is chaotic

INPUT:
    Question:{question}
    Reference Answer:{reference}
    Candidate Answer:{candidate}

OUTPUT:
    Please output ONLY a JSON object with 5 integer fields in the range [1,20], corresponding
    to the evaluation scores:
        {
        "correctness": <1-20>,
        "completeness": <1-20>,
        "relevance": <1-20>,
        "clarity": <1-20>,
        "reasoning": <1-20>
        }

SCORING INSTRUCTIONS:
- Read the reference answer carefully and identify ALL key points, details, and structure
- Compare the candidate answer systematically against the reference answer
- For each criterion, start with a conservative score and only increase if the candidate truly deserves it
- If the candidate answer is significantly shorter, less detailed, or less precise than the reference, reduce scores accordingly
- If the candidate answer contains information not in the reference (unless it's clearly relevant and accurate), consider reducing relevance score
- When scoring, ask yourself: "Does this candidate answer match the quality and completeness of the reference answer?" If not, reduce scores
- Average or mediocre answers should receive scores in the 8-15 range, not higher
- Only truly excellent answers that closely match the reference should receive 16-20 scores

REQUIREMENT:
    No explanation, no extra text, no formatting other than valid JSON. Be strict and conservative with your scores.
\end{judgepromptbox}

\section{Decisions after Removing Harness Cues}
\label{app:decision-probe}

Task scores show whether an agent succeeds with less external support. This
9B probe asks a narrower question: with the decision history held fixed,
does removing control information change the model's next action?
The four checkpoints are scored on the same reference states, rather than
on independently generated trajectories.

\subsection{Reference states and paired inputs}

We use archived $H_4$-trained evaluations under $H_4$ on
SWE-QA and SWE-QA-Pro. Candidate runs must be completed with \texttt{done},
have zero verifier failures and repairs, lexical F1 of at least 0.10,
reference-keyword recall of at least 0.25, and a code reference.
These are quality filters, not a guarantee that the answer is correct.
We match the recorded final answer to the prediction and resolve stored
tool observations. The $H_4$ training source contains 1,279 trajectory records covering
464 distinct questions. A post-hoc audit of the 128 probe questions
against the original question texts from all four source pools found
no exact matches after lowercasing and whitespace normalization.

With seed 6542, we select 64 distinct questions per benchmark. For each
trajectory we take the first, middle, and last tool decision and the final
answer, removing duplicated boundaries and examples above a 220,000-character
prefilter. This yields 432 candidate states. Keeping only states that fit
65,536 tokens in all 16 checkpoint--input combinations, without truncation,
leaves 395 states from 128 questions: 204 SWE-QA and 191 Pro states.
There are 319 tool decisions and 76 final-answer decisions; the late/final
subset contains 84 tool decisions and 76 answers.

Each state supplies four input-cue conditions, denoted $H_1$--$H_4$ in Table~\ref{tab:probe-matrix}. These labels refer to retained input cues here. $H_4$ keeps the original input;
$H_3$ removes the verification instruction block; $H_2$ also removes
the workflow instruction block; $H_1$ additionally removes the state
instruction and messages marked as harness-generated. The question, tool
schemas, tool observations, previous model messages, and target response
are unchanged. These are nested \emph{input-cue ablations}, not four reruns
of the executable harness. In particular, prior model history can retain
effects of the original harness.

\subsection{What the three metrics measure}

All scores use teacher forcing: the model sees the reference reasoning
before the action, and each predicted token is conditioned on the reference
token prefix. Thus the probe tests action selection \emph{given that
reasoning}, not the ability to generate it unaided. No tools are executed.

\begin{metricdefinition}{Stop/continue macro-F1}{395 decision states}
At the first action token, a top-1 prediction of the tool-call marker means
continue; any other token is counted as stop. The reference label is
continue for a tool action and stop for a final answer.
\[
 F_{\mathrm{stop/continue}}=\tfrac12(F_{1,\mathrm{stop}}+
 F_{1,\mathrm{continue}}),\qquad
 F_{1,c}=\frac{2TP_c}{2TP_c+FP_c+FN_c}.
\]
This is a binary marker diagnostic; a non-tool token need not be a valid
final answer.
\end{metricdefinition}

\begin{metricdefinition}{Late/final decision accuracy}{160 late/final states}
On the late/final subset, we report
\[
 A_{\mathrm{late/final}}=
 \frac{1}{|S_{\mathrm{lf}}|}
 \sum_{i\in S_{\mathrm{lf}}}\mathbf{1}[\hat c_i=c_i].
\]
This metric measures agreement with the reference continue/stop decision on
the late/final subset. There is no independent annotation of evidence
sufficiency. Because it uses a subset of the first metric's states, it is
not independent corroboration of control transfer.
\end{metricdefinition}

\begin{metricdefinition}{Next-tool accuracy}{319 tool decisions}
The first function name is correct only if every token in its name has
the correct teacher-forced top-1 prediction:
\[
 A_{\mathrm{tool}}=\frac{1}{319}\sum_i
 \mathbf{1}[\forall j,\hat t_{ij}=t_{ij}].
\]
The metric does not require correct arguments or successful tool execution.
\end{metricdefinition}

\subsection{Where control decisions improve}

Table~\ref{tab:probe-matrix} reports all 48 behavioral values. From
$H_4$-trained to $H_1$-targeted, stop/continue macro-F1 under $H_1$ input cues
rises from 64.43 to 87.79 and late/final decision accuracy from 62.50 to
83.13. Next-tool accuracy changes much less, from 96.24 to 97.18. The main
change is therefore in the conditional continue/stop prediction under
stripped inputs, rather than recognition of the reference tool name.

\begingroup
\appendixTableStyle
\normalsize
\renewcommand{\arraystretch}{1.16}
\setlength{\LTleft}{0pt}
\setlength{\LTright}{0pt plus 1fill}
\setlength{\LTpre}{6pt}
\setlength{\LTpost}{10pt}
\setlength{\LTcapwidth}{\linewidth}
\begin{longtable}{
  >{\small\raggedright\arraybackslash}p{\dimexpr0.32\linewidth-2\tabcolsep\relax}
  *{4}{>{\small\raggedleft\arraybackslash}p{\dimexpr0.17\linewidth-2\tabcolsep\relax}}
}
\caption{\textbf{Stopping decisions recover under $H_1$ inputs; tool-name prediction changes little.}
Teacher-forced scores (\%) on the same reference states. Columns change
input cues, not executable harnesses. Bold marks the best value within
each column and metric.}
\label{tab:probe-matrix}\\
\appendixTopRule
\rowcolor{appendixHeader}
& \multicolumn{4}{c}{\textbf{Input-cue condition}} \\
\rowcolor{appendixHeader}
\textbf{Checkpoint} & $\boldsymbol{H_1}$ & $\boldsymbol{H_2}$ & $\boldsymbol{H_3}$ & $\boldsymbol{H_4}$ \\
\midrule[0.45pt]
\endfirsthead
\caption[]{Behavioral probe results (continued).}\\
\appendixTopRule
\rowcolor{appendixHeader}
& \multicolumn{4}{c}{\textbf{Input-cue condition}} \\
\rowcolor{appendixHeader}
\textbf{Checkpoint} & $\boldsymbol{H_1}$ & $\boldsymbol{H_2}$ & $\boldsymbol{H_3}$ & $\boldsymbol{H_4}$ \\
\midrule[0.45pt]
\endhead
\midrule[0.35pt]
\multicolumn{5}{r}{\footnotesize Continued on the next page}\\
\endfoot
\appendixBottomRule
\endlastfoot

\rowcolor{appendixWash}
\multicolumn{5}{l}{\small\textcolor{appendixInk}{\textbf{Stop/continue macro-F1}}\quad ($N=395$)} \\*
$H_4$-trained & 64.43 & \textbf{94.31} & \textbf{95.75} & \textbf{95.71} \\*
$H_3$-targeted & 78.12 & 93.25 & 94.66 & 94.66 \\*
$H_2$-targeted & 83.95 & 92.57 & 94.31 & 94.31 \\*
$H_1$-targeted & \textbf{87.79} & 92.23 & 93.60 & 93.95 \\
\midrule[0.35pt]
\rowcolor{appendixWash}
\multicolumn{5}{l}{\small\textcolor{appendixInk}{\textbf{Late/final decision accuracy}}\quad ($N=160$)} \\*
$H_4$-trained & 62.50 & \textbf{92.50} & \textbf{94.38} & \textbf{94.38} \\*
$H_3$-targeted & 73.13 & 91.25 & 92.50 & 92.50 \\*
$H_2$-targeted & 78.75 & 90.00 & 91.88 & 91.88 \\*
$H_1$-targeted & \textbf{83.13} & 89.38 & 90.63 & 91.25 \\
\midrule[0.35pt]
\rowcolor{appendixWash}
\multicolumn{5}{l}{\small\textcolor{appendixInk}{\textbf{Next-tool accuracy}}\quad ($N=319$)} \\*
$H_4$-trained & 96.24 & 97.18 & \textbf{95.92} & 95.92 \\*
$H_3$-targeted & 96.55 & 97.49 & 95.61 & \textbf{96.24} \\*
$H_2$-targeted & 96.55 & \textbf{97.81} & 95.61 & 95.61 \\*
$H_1$-targeted & \textbf{97.18} & \textbf{97.81} & 95.61 & 95.92 \\
\end{longtable}
\arrayrulecolor{black}
\endgroup

For metric $m$, define the descriptive behavioral gap as
$G_m(M)=m(M,H_4)-m(M,H_1)$.
The stop/continue gap contracts from 31.28 to 6.16 percentage points;
the late/final gap contracts from 31.88 to 8.12 points.
This is not cost-free: $H_4$-input F1 falls from 95.71 to 93.95,
and late/final accuracy from 94.38 to 91.25. The matrix separates
weak-input improvement from deterioration under full cues rather than
treating a smaller gap alone as successful transfer.

\par\noindent\begin{minipage}{\linewidth}
\section{Workflow Recovery Case}
\label{app:trajectory-analysis}

Figure~\ref{fig:paired-workflow-case} compares the $H_4$-trained and
$H_3$-targeted checkpoints on the same task, both deployed under $H_2$.
The pair illustrates recovery and stopping after workflow control is
removed; it is not an aggregate effect or proof that the final answer
is grounded.

\definecolor{haBlue}{HTML}{2F5D7C}
\definecolor{haRed}{HTML}{A53A32}
\definecolor{haGreen}{HTML}{2F6B55}
\definecolor{haMid}{HTML}{6F7377}
\definecolor{haRule}{HTML}{B8BDC2}
\definecolor{haPanel}{HTML}{F5F6F7}
\definecolor{haBlueWash}{HTML}{EDF3F7}
\definecolor{haRedWash}{HTML}{FAEFED}
\definecolor{haGreenWash}{HTML}{EBF4F0}
\begin{figure}[H]
\centering
\resizebox{\linewidth}{!}{%
\begin{tikzpicture}[x=1cm,y=1cm,font=\normalsize]
\node[anchor=north west,text width=16.7cm,inner sep=0pt] at (0,0) {%
  \textbf{Same task under $H_2$:}
  how does \texttt{ensure\_eol()} modify the output buffer?};
\fill[haPanel] (0,-0.78) rectangle (8.08,-1.95);
\draw[haRule,line width=0.45pt] (0,-0.78) rectangle (8.08,-12.05);
\fill[haPanel] (8.62,-0.78) rectangle (16.70,-1.95);
\draw[haRule,line width=0.45pt] (8.62,-0.78) rectangle (16.70,-12.05);
\node[anchor=west,font=\bfseries\small] at (0.30,-1.20)
  {A\quad Before $H_3$-targeted training};
\node[anchor=west,text=haMid,font=\footnotesize] at (0.30,-1.66)
  {$H_4$-trained checkpoint};
\node[anchor=west,font=\bfseries\small] at (8.92,-1.20)
  {B\quad After $H_3$-targeted training};
\node[anchor=west,text=haMid,font=\footnotesize] at (8.92,-1.66)
  {$H_3$-targeted checkpoint};
\foreach \x in {0.56,9.18}{
  \draw[haRule,line width=0.8pt] (\x,-2.70) -- (\x,-10.05);
  \foreach \y/\n in {-2.70/1,-4.25/2,-5.80/3,-7.85/4,-9.95/5}{
    \fill[white] (\x,\y) circle (0.15);
    \draw[haRule,line width=0.7pt] (\x,\y) circle (0.15);
    \node[font=\scriptsize,text=haMid] at (\x,\y) {\n};
  }
}
\node[anchor=north west,text width=6.75cm,align=left,inner sep=0pt]
  at (0.96,-2.47) {Plans to locate the translator, footnote visitor,
  line terminator, and output buffer.};
\node[anchor=north west,text width=6.75cm,align=left,inner sep=0pt]
  at (0.96,-4.02) {Uses \texttt{Glob/Grep}, then reads candidate writer files.};
\filldraw[fill=haRedWash,draw=haRed,line width=0.7pt]
  (0.88,-5.20) rectangle (7.78,-8.85);
\node[anchor=north west,text width=6.38cm,align=left,inner sep=0pt]
  at (1.14,-5.53) {\textcolor{haRed}{\bfseries Unresolved subgoal}\\[4pt]
  \lq\lq I need to find the parent class \texttt{Translator} implementation.''};
\node[anchor=north west,text width=6.38cm,align=left,inner sep=0pt]
  at (1.14,-7.57) {The same goal is restated across \mbox{steps 3--8}.
  Similar \texttt{Grep/Read} calls add no new evidence.};
\node[anchor=north west,text width=6.75cm,align=left,inner sep=0pt]
  at (0.96,-9.53) {No strategy switch and no final answer.
  The loop guard terminates the run.};
\node[anchor=north west,text width=6.75cm,align=left,inner sep=0pt]
  at (9.58,-2.47) {Starts from the same localization plan.};
\node[anchor=north west,text width=6.75cm,align=left,inner sep=0pt]
  at (9.58,-4.02) {Detects that repository search is unavailable.};
\filldraw[fill=haBlueWash,draw=haBlue,line width=0.7pt]
  (9.50,-5.20) rectangle (16.40,-7.25);
\node[anchor=north west,text width=6.38cm,align=left,inner sep=0pt]
  at (9.76,-5.53) {\textcolor{haBlue}{\bfseries Recovery decision}\\[4pt]
  Switches to shell search and direct source inspection.};
\node[anchor=north west,text width=6.75cm,align=left,inner sep=0pt]
  at (9.58,-7.57) {Locates the override and traces its call to
  \texttt{super().visit\_footnote(node)}.};
\filldraw[fill=haGreenWash,draw=haGreen,line width=0.7pt]
  (9.50,-9.20) rectangle (16.40,-11.15);
\node[anchor=north west,text width=6.38cm,align=left,inner sep=0pt]
  at (9.76,-9.53) {\textcolor{haGreen}{\bfseries Stopping decision}\\[4pt]
  Answers using the located helper despite incomplete parent inspection.};
\fill[haRedWash] (0.24,-11.37) rectangle (7.84,-11.82);
\node[anchor=west,font=\bfseries\footnotesize,text=haRed] at (0.42,-11.60)
  {LOOP DETECTED};
\node[anchor=east,font=\scriptsize,text=haMid] at (7.65,-11.60)
  {8 steps \quad 20 calls \quad no answer};
\fill[haGreenWash] (8.86,-11.37) rectangle (16.46,-11.82);
\node[anchor=west,font=\bfseries\footnotesize,text=haGreen] at (9.04,-11.60)
  {DONE};
\node[anchor=east,font=\scriptsize,text=haMid] at (16.27,-11.60)
  {11 steps \quad 20 calls \quad answer returned};
\end{tikzpicture}%
}
\caption{\textbf{Workflow recovery and stopping under $H_2$.}
Both 9B runs use \texttt{sphinx-00045} at commit \texttt{6c9e320} and the
same four-tool interface, with external workflow control absent.
The question concerns the output buffer before the parent footnote visitor.
The $H_4$-trained run repeats an unresolved search; the $H_3$-targeted
run changes strategy and returns an answer after the same total of 20 tool calls.
Text is abridged from the recorded reasoning.
Both runs stop below their active step limits, and no displayed response is
length-truncated. The returned answer cites a same-named helper in
\texttt{texinfo.py} without verifying the inherited \texttt{docutils}
implementation. The case therefore documents recovery and stopping behavior;
it does not establish that the final answer is fully grounded.}
\label{fig:paired-workflow-case}
\end{figure}
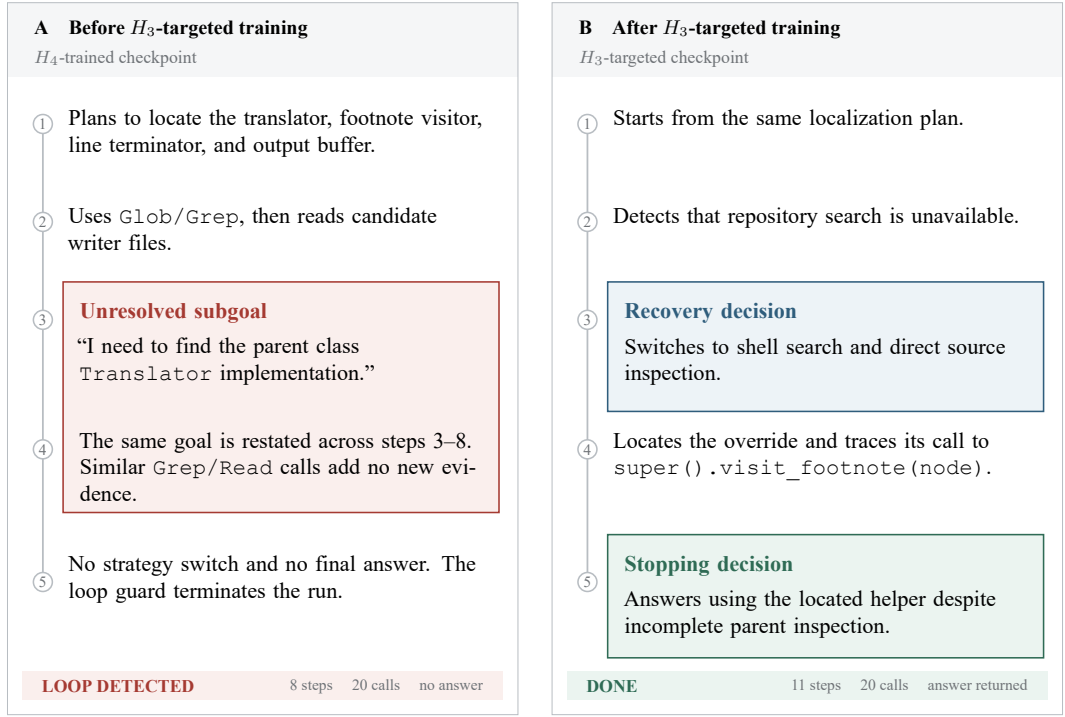

\end{minipage}\par



\end{document}